\documentclass[hidelinks,conference]{IEEEtran}
\usepackage{cite}
\usepackage{amsmath,amssymb,amsfonts}
\usepackage{algorithmic}
\usepackage{graphicx}
\usepackage{textcomp}
\usepackage{xcolor}
\usepackage{pgfplots}
\usepackage{booktabs}
\usepackage{float}
\usepackage{url}
\usepackage{hyperref}

\usepackage{caption}
\pgfplotsset{compat=1.18}

\usepackage{xspace}
\newcommand{\ie}{\emph{i.e.,}\xspace}
\newcommand{\eg}{\emph{e.g.,}\xspace}

\newcommand{\etal}{\emph{et~al.}\xspace}

\usepackage{pifont}

\usepackage{tcolorbox}
\newcommand{\highlight}[1]{\begin{tcolorbox}\textit{#1}\end{tcolorbox}}

\usepackage{fontawesome}
\newcommand{\totalpapers}{34}
\newcommand{\papercount}[1]{\textit{\textbf{#1} out of~\totalpapers~papers}}

\def\BibTeX{{\rm B\kern-.05em{\sc i\kern-.025em b}\kern-.08em
    T\kern-.1667em\lower.7ex\hbox{E}\kern-.125emX}}
    
\usepackage{enumitem}
\newlist{questions}{enumerate}{2}
\setlist[questions,1]{label=\textbf{RQ.},ref=RQ}
\setlist[questions,2]{label=(\alph*),ref=\thequestionsi(\alph*)}

\usepackage{multibib}
\newcites{P}{Primary Studies}

\usepackage{balance}

\usepackage{listings}
\begin{document}

\title{The Two Faces of AI in Green Mobile Computing: A Literature Review}

\author{

\author{
\IEEEauthorblockN{Wander Siemers\IEEEauthorrefmark{1},
June Sallou\IEEEauthorrefmark{1}, Lu\'is Cruz\IEEEauthorrefmark{1}}
\IEEEauthorblockA{\IEEEauthorrefmark{1}Delft University of Technology, The Netherlands - {wandersiemers@me.com,  \{ j.sallou, l.cruz \}@tudelft.nl}}
}
}

\maketitle

\begin{abstract}

Artificial intelligence is bringing ever new functionalities to the realm of mobile devices that are now considered essential (e.g., camera and voice assistants, recommender systems). Yet, operating artificial intelligence takes up a substantial amount of energy.
However, artificial intelligence is also being used to enable more energy-efficient solutions for mobile systems. 
Hence, artificial intelligence has two faces in that regard, it is both a key enabler of desired (efficient) mobile functionalities and a major power draw on these devices, playing a part in both the solution and the problem. 
In this paper, we present a review of the literature of the past decade on the usage of artificial intelligence within the realm of green mobile computing.
From the analysis of 34 papers, we highlight the emerging patterns and map the field into 13 main topics that are summarized in details. 
%

Our results showcase that the field is slowly increasing in the past years, more specifically, since 2019. Regarding the double impact AI has on the mobile energy consumption, the energy consumption of AI-based mobile systems is under-studied in comparison to the usage of AI for energy-efficient mobile computing, and we argue for more exploratory studies in that direction.
We observe that although most studies are framed as solution papers (94\%), the large majority do not make those solutions publicly available to the community.  
Moreover, we also show that most contributions are purely academic (\papercount{28}) and that we need to promote the involvement of the mobile software industry in this field.


\end{abstract}


\section{Introduction}

Artificial intelligence (AI) is bringing ever more new functionalities to the realm of mobile devices that are now considered essential (e.g., camera and voice assistants, recommender systems).
Users have increasing expectations of the processing power and capabilities of their mobile devices. Contemporary smartphones have highly advanced image processing systems, smart integrated assistants, and offer gigabit speeds over their radios. All of these features, and many more, are enabled by artificial intelligence \cite{ignatov2021fast, schuster2010speech},\citeP{memon2019artificial}.

Yet, operating artificial intelligence takes up a substantial amount of energy. Modern artificial intelligence techniques, such as deep learning, can have very high energy consumption, both on dedicated servers \cite{zhu2018benchmarking} and on mobile devices \cite{liu2019performance}. 


Beyond the realm of artificial intelligence, mobile computing has long been concerned with energy efficiency due to the limited power capacity of smartphones~\cite{cruz2017performance,cruz2019catalog}.
Less obviously, artificial intelligence techniques themselves can also be used to reduce mobile energy consumption. For example, by optimizing data transmission~\citeP{memon2019artificial}, or location services~\citeP{donohoo2013context}. 
Hence, artificial intelligence has two faces in this problem: it is both 1) a key enabler of desired (efficient) mobile features and 2) a major power draw on these devices, playing a part in both the solution and the problem.
In this paper, we provide a comprehensive overview of both of these aspects of mobile energy use and artificial intelligence by reviewing the associated literature. The goal of this review is to understand the characteristics of the literature on mobile energy consumption involving artificial intelligence.

Our literature review yields 34 papers from 2013 until late 2022. We identify and pinpoint thirteen different topics being addressed by the literature.
Our results showcase a growing interest in the intersection between AI and Mobile Computing Energy since 2019. Most studies revolve around solution papers (\papercount{32}) and only \papercount{6} display the participation of authors with an industry affiliation.
We argue that it is quintessential that contributions in this field come with a replication package and that proposed solutions are made available to the public.
Finally, although topics such as Approximate Computing and Benchmarking have been marginally covered  by the literature (\papercount{2}), we expect them to be relevant to the challenges posed in this field.

The contributions of this paper are three-fold:

\begin{itemize}
    \item An analysis of the field of AI in Green Mobile Computing, covering publications per year, study type, industry involvement, level of study, and tool provision. 
    \item A mapping of the field into different topics, with the respective summary of existing contributions.
    \item A replication package that provides all the collected data for each paper, that can be used in future reviews.\footnote{Replication package: \url{https://zenodo.org/record/8172245}} 
\end{itemize}

The remainder of this paper is structured as follows. We describe the detailed methodology in Section \ref{sec:methods}. Following this methodology, we present our results in section \ref{sec:results}. We then discuss these results and their related impacts in the research community in Section~\ref{sec:discussion}.
The threats to validity of our study can be found in Section~\ref{sec:threats}. We then treat related work in section \ref{sec:related} and discuss how our work differs from those studies. Lastly, we highlight the conclusions of the literature review in section~\ref{sec:conclusion}.

\begin{figure*}[!ht]
    \centering
    \vspace{2ex}
    \includegraphics[width=1.0\textwidth]{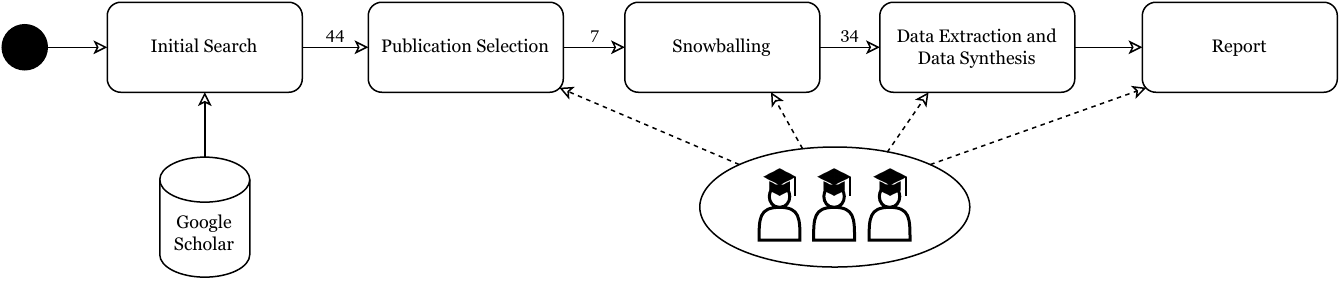}
    \caption{Process overview of the systematic literature review.\label{fig:process}}
    \vspace{-1em}
\end{figure*}

\section{Methodology} \label{sec:methods}
In this section, we present the methodology we rigorously followed while carrying out this review. We follow the guidelines for conducting literature reviews in software engineering research presented by several dedicated publications~\cite{Kitchenham2004, wohlin2014guidelines, petersen2015guidelines, mayring2004qualitative}.


\subsection{Research Goal and Question} \label{sec:rq}
The goal of this review is to understand the characteristics of the literature on mobile energy consumption involving artificial intelligence.
It can be directly translated into a research question (RQ), which states as follows:
\begin{questions}
    \item \textit{What are the characteristics of the state-of-the-art research regarding Artificial Intelligence in Green Mobile Software?}\label{rq2}
\end{questions}

We are interested in determining the trends in publishing in this field (e.g., the evolution of the number of publications, the study environment, and tool provision), as well as the trends in the studies themselves (e.g., topics, types of AI usage).

\subsection{Research Process}

The different steps of the research process are illustrated in Figure~\ref{fig:process}. It begins with an initial search conducted through an automatic query on the Google Scholar database, which is then augmented by a bidirectional snowballing process. The subsequent sections provide a detailed description of each step.


\subsubsection{Initial Search} 
To gather an initial set of publications, we define a search query that is executed on the \textit{Google Scholar} database. The query states as follows:
\vspace{-3em}
\begin{lstlisting}[escapechar=@,  language=sql, caption={Search query}, label={list:query}, numbers=none, abovecaptionskip=3em, basicstyle=\small	\ttfamily, abovecaptionskip=3em]
("AI-Based" OR "machine learning" OR 
"Artificial Intelligence") AND "mobile" 
AND ("energy" OR "efficient" OR "green")
\end{lstlisting}

\vspace{-0.5em} 
The query is designed to retrieve publications whose titles match keywords related to the concepts of AI (\ie~\textit{AI-Based}, or \textit{machine learning}, or \textit{Artificial Intelligence}), mobile computing (\ie~\textit{mobile}), and energy efficiency (\ie~\textit{energy} \textit{efficient}, or \textit{green}). We focus on titles to target literature whose main topics deal with green mobile computing and AI, restricting the inital pool of results but attempting to increase the relevance of the matches.
We perform this search using the \textit{Publish or Perish}\footnote{Publish or Perish software, \url{https://harzing.com/resources/publish-or-perish}} software, searching on \textit{Google Scholar} and matching only on title keywords (\ie by using the ``title words'' option). We select \textit{Google Scholar} for several reasons: (i) it indexes results from various libraries, allowing us to perform only a single search, (ii) as reported in the set of guidelines by Wohlin~\cite{wohlin2014guidelines}, the use of such an indexer is a suitable choice for identifying the initial set of literature for snowballing processes, (iii) the query results can be automatically extracted from the indexer using Publish or Perish.

The initial search was performed on the October 19th, 2022, and provided 44 results using Publish or Perish\footnote{Performing the query with Publish or Perish is the equivalent of the Advanced Google Scholar Search with the query ``\textit{allintitle: mobile energy OR efficient "AI Based" OR "machine learning" OR "Artificial Intelligence”}''.}.

\subsubsection{Publication Selection}

The initial set of publications undergoes a selection process based on predefined selection criteria. Inclusion (\textbf{I}) and exclusion (\textbf{E}) criteria are established to ensure the included papers in the primary study set are not only relevant to our research question but also of high quality.

\begin{itemize}
\item \textbf{I-1.} The study regards mobile devices (smartphones/tablets)
\item \textbf{I-2.} The study regards energy consumption
\item \textbf{I-3.} The study regards artificial intelligence\footnote{We have used a broad interpretation of the concept `AI', including papers proposing simple techniques like regression.}: either by treating how AI can be used to reduce mobile energy consumption or by treating the energy consumption of mobile AI itself
\item \textbf{I-4.} The study regards the software level

\item \textbf{E-1.} The study is not written in English
\item \textbf{E-2.} The study is not accessible
\item \textbf{E-3.} The study is not peer-reviewed
\item \textbf{E-4.} The study is in the form of citations, patents, editorials, tutorials, books, extended abstracts, thesis, etc.
\item \textbf{E-5.} The study is not a primary study, such as a review paper. 
\item \textbf{E-6.} The study was published before 2012
\end{itemize}

The first three inclusion criteria ensure that the selected papers deal with the topic of artificial intelligence in mobile energy consumption (\textbf{I-1} to \textbf{I-3}). The fourth criterion (\textbf{I-4}) guarantees that the studies focus on the software level. The goal is to exclude papers that focus exclusively on hardware-specific approaches (\eg~involving peculiar hardware components to improve the mobile energy consumption while using AI). 
For the exclusion criteria, the first two criteria (\textbf{E-1} and \textbf{E-2}) assure that we can extract data from the papers. The two following criteria (\textbf{E-3} to \textbf{E-5}) make sure the papers constitute scientific primary studies.
We include an additional criterion, \textbf{E-6}, to exclude papers published before 2012. This decision is based on the rapid changes in both AI and mobile technology over the past decade. Hence, we believe that studies before that date are not particularly relevant, especially when considering the relatively new trend of energy efficiency in the topic.


The publications in the initial set are distributed among the three authors for assessment based on the selection criteria. Each author independently evaluates the assigned publications using adaptive reading depth~\cite{petersen2015guidelines}. Moreover, regular meetings are conducted to facilitate discussions regarding the selection process and to minimize any potential personal biases.

After applying the in/exclusion criteria to the 44 retrieved papers of the initial set, we identify 7 primary studies.

\subsubsection{Snowballing}
To address the limitations of our initial query and ensure a comprehensive representation of the literature on AI and mobile computing energy, we employ a bidirectional snowballing technique. This approach aims to supplement the primary studies by retrieving papers that may have been missed by the title-only search query. Following the guidelines presented by Wohlin~\cite{wohlin2014guidelines}, we conduct a single iteration of snowballing, encompassing both backward and forward passes for the papers in the initial set. During this process, we thoroughly examine all studies that either cite (forward snowballing) or are cited (backward snowballing) by the primary studies already included in our analysis.

To maintain objectivity and rigour, the three authors independently explore different primary studies, identifying additional studies that align with the predefined selection criteria for inclusion. Throughout this exploration, any doubts or disagreements that arise during the assessment are addressed through discussions among the authors. These discussions serve as a mechanism to mitigate subjective biases and ensure a collective resolution for all assessments.

Finally, the snowballing process results in the addition of 27 new primary studies, leading to a total of $34$ primary studies, which are examined in this literature review.

\subsubsection{Data Extraction}
Once the final set of primary studies is completed, we proceed to the systematic data extraction step to answer our research question (cf.~\textbf{\ref{rq2}} in Section~\ref{sec:rq}). 

To establish the specific data fields to be extracted from the primary studies, the authors independently review them initially and annotate the relevant characteristics that address the research question. These characteristics are subsequently subject to thorough discussions and refinement through open coding techniques~\cite{mayring2004qualitative, maurer2008card}. By engaging in open coding, the authors ensure a more precise identification and categorization of the study attributes. This iterative process allows for the final determination of the fields to be utilized during the subsequent data extraction phase.

Once the data fields are finalized, the authors re-examine the papers meticulously, rigorously analysing them to extract the data corresponding to the identified fields of interest. This data extraction process involves a comprehensive and systematic approach, ensuring that the desired information is accurately captured from the primary studies.

The fields used for the data extraction are the following:
\begin{itemize}
    \item \textbf{Publication Year} 
    \item \textbf{Study Type:} The type of study the paper is presenting: either a \textit{position} on AI in Green Mobile Software, a \textit{solution} to tackle an issue on the topic, or an \textit{observational} study; 
    \item \textbf{Category of AI Role:} The role AI has regarding Green Mobile Computing. It can either be the use of AI for improving the energy efficiency of mobile computing, or the study of energy consumption of AI-based mobile systems; 
    \item \textbf{Topic:} The topic the primary study is focusing on. For instance, context adaptation, in which the mobile software execution is readjusted according to the context, to improve the energy efficiency;
    \item \textbf{Level of Study:} It corresponds to the scale at which the mobile software is studied (either at the level of the \textit{device}, or of the \textit{system});
    \item \textbf{Industry Involvement:} The involvement of industry in the authoring of the study, which can be either exclusively academic, exclusively industrial or a mix;
    \item \textbf{Tool Provision:} The availability of the tool(s) to handle AI in Green Mobile Computing presented in the study (if applicable).
\end{itemize}

\subsubsection{Data Analysis}
Along the whole process, the authors discussed the emerging codes generated during data extraction to ensure that they are congruent with each other, meet the research objective and answer the question being addressed.

Regarding the field of \textbf{Topic}, the approach of \textit{open coding} was used to group the different keywords into a coherent hierarchical structure~\cite{mayring2004qualitative, maurer2008card}. 
For the rest of the fields, the keywords were already pre-set in advance. In the case of \textbf{Study Type}, the different options were \textit{position}, \textit{observational}, and \textit{solution}. As for \textbf{Category of AI Role}, they were \textit{AI4E}, to translate the use of AI to make the mobile software more energy efficient, and \textit{EofAI}, for the fact that AI was involved in the design of the mobile software itself, and was studied regarding energy efficiency.
The \textbf{Level of Study} field was given the keywords \textit{Device} and \textit{System}, to represent the scale at which the study was focusing.
The \textbf{Industry Involvement} was based on the terms \textit{Academic} for academic authorship only, \textit{Industrial} for industrial authorship only, and \textit{Mix} for mixed authorship.
In the case of \textbf{Tool Provision}, the options were reduced to either \textit{Yes} and \textit{No}.
Finally, for the \textbf{Publication Year}, the year was directly extracted from the publication date.

\section{Results} \label{sec:results}
In this section, we present the results collected with our systematic literature review on the roles of artificial intelligence in mobile energy efficiency.

\subsection{Publication Years}
\label{sec:year}
\begin{figure}[!ht]
    \centering
    \includegraphics[width=1.0\linewidth]{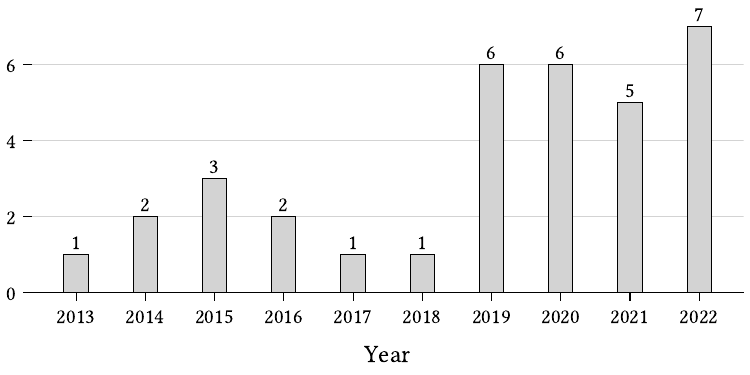}
    \vspace*{-1em}
    \caption{Publication of papers per year.\label{fig:plot_year}}
    \vspace{-1.5em}
\end{figure}

For the past decade, we can observe a trend of an increase in the number of papers dealing with Artificial Intelligence with respect to mobile energy consumption. Figure~\ref{fig:plot_year} shows an apparent increase from year 2019, going from approximatively 2 papers per year before 2019, to 6 papers being published per year after 2019. As the number of papers remains small, that increase needs to be interpreted with caution. 
Furthermore, it should be noted that the results for the year 2022 do not correspond to the full year and may not be representative of the actual research output, as the automated initial search was executed in October 2022.
\vspace{-1em}

\highlight{\textbf{Publication Trends:}
The research topic of artificial intelligence in green mobile software has been gaining in popularity since 2019, and appears to be on track to continue doing so.}

\subsection{Types of Study} \label{sec:type}
The literature on the mobile energy consumption related to artificial intelligence from the last decade is predominantly composed of \textit{solution} papers with \papercount{32}. We can notice that, in comparison, there are only 2 \textit{observational} papers and no \textit{position} papers.
Note that study types are mutually exclusive, i.e., a single paper has only one study type.

\highlight{\textbf{Type of Study:}
The majority of the literature consists of studies proposing solutions, with a very small number of observational studies.}

\subsection{Category of AI Role}\label{sec:category}
\begin{figure}[!ht]
    \centering
    \includegraphics[width=0.65\linewidth]{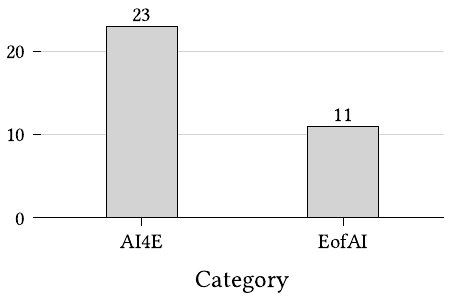}
    \caption{Distribution of papers per category (\ie~AI4E or EofAI).\label{fig:plot_category}}
\end{figure}

Around 68\% of the papers (\ie~\papercount{23}) deal with the use of artificial intelligence to tackle the mobile energy consumption (cf.~\textit{AI4E} in Figure~\ref{fig:plot_category}), while the rest (\ie~\papercount{11}) studies the energy consumption of the use of artificial intelligence in mobile devices (cf.~\textit{EofAI} in Figure~\ref{fig:plot_category}). 

\highlight{\textbf{Category of AI Role:}
The use of AI to make mobile software more energy-efficient is being studied twice as much as the energy consumption of AI-based mobile software.}

\subsection{Topics}\label{sec:topic}
We identified 13 different topics that are covered by the literature on the use of artificial intelligence in mobile in the context of energy efficiency.  
Figure~\ref{fig:plot_topic} presents the distribution of the studies according to the topics.
The most popular topic is \textit{Offloading}, addressed by 9 papers, followed by \textit{Context Adaptation} (8), and \textit{Federated Learning} (7).
Since papers are not exclusive to a single topic, these top-3 topics alone cover 70\% of the papers in this review.
Below, we pinpoint each topic with a short summary and the respective number of publications.

\textbf{Offloading:} \papercount{9}. 
Approach to delegate the execution of 
a resource-intensive task from a lower-powered device, often a mobile device, to a different, usually more powerful, device or service~\cite{somula2019research}.
Frequently, a mobile device offloads to the cloud, in which case the technique is also called \textit{Mobile Cloud Computing} (MCC) ~\citeP{zhang2015offloading, nawrocki2019adaptable, akki2020energy}.
The papers covering this topic study how to optimise the offloading process by employing AI. For instance, 
Markov Decision Processes (MDPs) are used to optimize offloading schedules in \citeP{chen2018performance, zhang2015offloading}. The objective of using these MDPs is to minimize the long-term cost, in which energy expended is included. Both traditional value iteration \citeP{zhang2015offloading} and approaches \citeP{chen2018performance} using a Deep Q-Network (DQN) (first proposed in \citeP{mnih2015human}) have been used to determine policies efficiently. In particular, Zhang~\etal~\citeP{zhang2015offloading} propose to tackle the problem of intermittence in connection, when the user mobile is out of communication range with the offloading {cloud) system.

\textbf{Context Adaptation:} \papercount{8}. The goal of the primary studies involving the topic is to improve the energy efficiency of mobile software by readjusting its execution according to the context.
Some papers are using the user of the mobile device as their reference for the definition of the context, while others deal with the device itself as the reference for context of execution.

For instance, Machidon~\etal~\citeP{machidon2022context} propose to optimise the resolution of video depending on (user) context. The authors observe that the personality traits of the user and the user's motion affect the perception of mobile video. The authors then use this observation to build a predictor for the desired resolution of a mobile video. Both a Random Forest and Mean Regressor are applied, with the Random Forest being the most accurate. However, the energy saved is not quantified, and importantly, the energy use of the predictor itself is not measured.

Regarding the mobile context perspective, Donohoo~\etal~\citeP{donohoo2013context} exploit the spatio-temporal and device context and use AI to predict and adapt device wireless data and location interface configurations so that energy consumption in mobile devices is optimised.
Nawrock~\etal~~\citeP{Nawrocki2020May} focus on the context of the device in a heterogeneous environment  to provide users with customized recommendations of products or services through the use of recommender systems (based on the use of AI).

\textbf{Federated Learning:} \papercount{7}. The application of AI to train a model across several decentralised devices with their own local subset of data, without the necessity of sharing data to all the involved devices.

The associated papers are looking at improving the communication and training execution strategies to improve the overall energy efficiency of such mobile software.

For example, a custom distributed learning system is provided by Deng~\etal~\citeP{deng2022making}. The authors propose multiple algorithms to schedule training efficiently while allowing setting limits on the energy consumption of the clients. It reaches higher accuracy scores than its baselines.

Shahidinejad~\etal~\citeP{shahidinejad2021context} considers a network full of devices, in which offloading decisions are interdependent. Each device learns using Deep Reinforcement Learning and shares its results with peers. The authors report a modest (2\%) energy consumption improvement compared to not offloading.

\textbf{Accuracy-Energy Trade-Off:} \papercount{5}.
The highest accuracy in AI-based software can lead to high energy consumption. Trading off some accuracy for better energy performances is a means to make mobile software greener.

For instance, Zheng~\etal~\citeP{zheng2022exploring} identify the logical trade-off between energy use and accuracy applied to Federated Learning: selecting the clients of the network with the largest data sets, the accuracy is increased, but the energy use grows accordingly since more training time and computation are needed on a larger data set.
The authors use POMPDP, a deep reinforcement learning technique, to optimize the ratio between training accuracy and energy consumption. They report a 51.8\% improvement in the ratio between accuracy and energy consumption.

Other application domains are explored, such as cellular networks.
In that direction, Ruiz~\etal~\citeP{Ruiz-Guirola2021Feb} use Discontinuous Reception (DRX), an energy-saving technique applied in 3G and 4G networks that turns off cellular hardware regularly to reduce energy consumption. The authors consider DRX on voice communications traffic. Using the fact that human speech contains periods of silence, the authors propose employing a Gaussian Process (GP) to optimize the duration for which cellular radios can remain off without interrupting speech. Energy savings of up to 30\% can be achieved compared to basic DRX schemes.

\textbf{Scheduling:} \papercount{5}. The scheduling, when the execution of a task is performed, can impact the energy efficiency of the mobile software.

Multiple solutions for learning optimal training schedules exist. Deep Q-Networks (DQN) can be used to learn \textit{resource budgets} for mobile clients \citeP{anh2019efficient, hieu2022deep} to deal with the constraints of mobile devices. This solution does not require any advance knowledge of network dynamics. However, DQN can suffer from a problem called \textit{over-optimistic value estimation} because it uses the same Q-value for selecting and evaluating an action \cite{van2016deep}. Therefore, Anh~\etal~\citeP{anh2019efficient} use a variant of DQN, called DQNN, which uses two neural networks to circumvent this problem. Huang~\etal~\citeP{huang2022worker} propose a very similar approach with two neural networks but runs it on the client instead of the server.

\textbf{Monitoring:} \papercount{4}.
Covering monitoring approaches to study the energy efficiency of mobile software.

Papers provide solutions to profile of energy consumption during runtime. 

In a first study, Aggarwal~\etal~\citeP{Aggarwal2014} rely on dynamic analysis of test cases to estimate the fluctuation of power use caused by modifications in the software. More specifically, they use the traces of system calls during application execution to gather metrics while software is undergoing a use case test. In a second paper, Aggarwal~\etal~\citeP{aggarwal2015greenadvisor} provide a tool called \textit{GreenAdvisor} to help mobile software practitioner with analysing the energy profile of their code, and to predict the impact of code changes on that energy profile.

With the study made by Pandey~\etal~\citeP{pandey2021here}, the different tasks executed in a network of mobile devices are profiled with regard to resource utilisation. AI is then used to analyse the profiles and to perform the best match between task to be executed and devices in order to improve the energy efficiency of the tasks.

Finally, Wang~\etal~\citeP{wang2020energy} address the energy profiling of mobile software involving augmented reality. The authors measure the energy consumption of a mobile Convolutional Neural Network (CNN) for computer vision and compare it to the energy consumption of an offloaded version of the model. They conclude that some light models can be run on recently released smartphones with low latency. Latency might even be lower than offloading the model to a remote server.

\textbf{Deployment:} \papercount{2}.
Dealing with how the mobile software is being deployed. 

Li~\etal~\citeP{li2020efficient} propose to employ AI to support intelligent network resource management strategies in mobile cloud computing to optimise the resource allocation.

In contrast, Tang~\etal~\cite{tang2014energy} address tail energy in their study. Tail energy is the energy that a cellular interface uses after data transmission has ended because it remains in a high-power state. This accounts for a significant fraction of total data transmission energy. The pattern of data transmission may be learned by artificial intelligence to allow the cellular interface to enter a low-power state sooner. Instead of training a model on-device, which might suffer from high resource use and the cold start problem, the system in \cite{tang2014energy} uses a client-server architecture. Mobile devices, therefore, send transmission records to a central server which trains an MLP classifier with two classes: 1) \textit{can be delayed} and 2) \textit{should be transmitted now}. Energy usage can be reduced by 20\% compared to always immediately sending requests.

\textbf{Other:} \papercount{6}.
Studies addressing a relevant topic with only a single publication in total: Benchmarking~\citeP{szabo2021machine}, Approximate Computing~\citeP{machidon2022context}, Model Design~\citeP{bhattacharya2016sparsification}, Recommenders~\citeP{nawrocki2016learning}, Energy Measurement~\citeP{mcintosh2019can}, and Resource Management~\citeP{li2020efficient}.

\begin{figure}[!ht]
    \centering
    \includegraphics[width=1.0\linewidth]{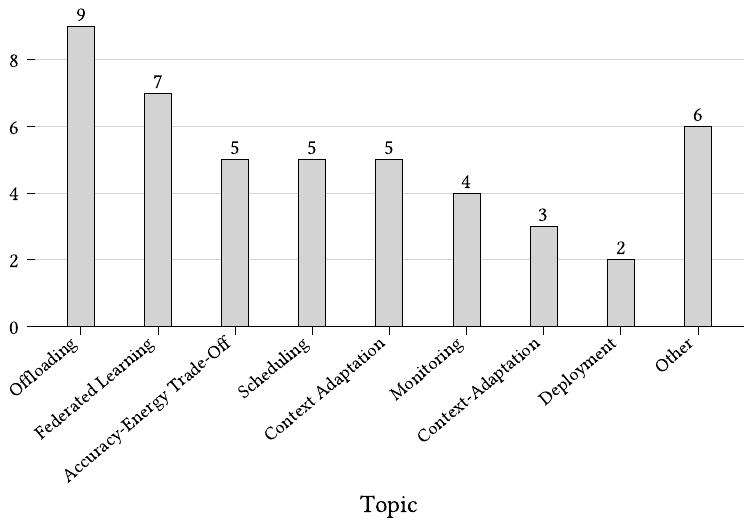}
    \caption{Topics addressed in the studies.}
    \label{fig:plot_topic}
\end{figure}

\highlight{\textbf{Topics:}
The literature on AI in Green Mobile Software covers 13 main topics. It focuses mainly on Offloading, Context Adaptation and Federated Learning, representing 70\% of the papers. In contrast, the following topics are currently under-explored: Benchmark, Approximate Computing, Model Design, Recommenders, Energy Measurement, and Resource Management.}

\subsection{Level of Study}\label{sec:level}
The mobile energy consumption can be addressed at different levels depending on the context of study. The majority of the papers (\papercount{22}) focus their studies at the system level. Rather than the energy consumption of a single mobile device, it is the consumption of a set of devices in a more complex setting, such as a network, and related tasks (\eg~transmission of data). The rest of the papers, around a third of them (\papercount{12}), tackle energy-consuming (AI) tasks executed on a single mobile device.

\highlight{\textbf{Level of Study:}
Approximately two thirds of the literature focus on Green Mobile software at the system-level, compared to the device-level. More attention is given to the topic in the context of a network of mobile devices.}

\subsection{Industry Involvement}\label{sec:industry}
Regarding the industry involvement in the publications on artificial intelligence in mobile energy consumption (cf.~Section~\ref{sec:methods}), an overview of the authorship of the primary studies is rendered in Figure~\ref{fig:plot_industry}.

From the figure, we can notice that the vast majority of the papers are authored by academic researchers (\papercount{28}), while 6 papers have authors being a mix of researchers with an academic and industrial background (cf.~\textit{Mix} in Figure~\ref{fig:plot_industry}). Additionally, we observe that there is no publication being exclusively authored by industrial researchers.

\begin{figure}[!ht]
    \centering
    \includegraphics[width=0.75\linewidth]{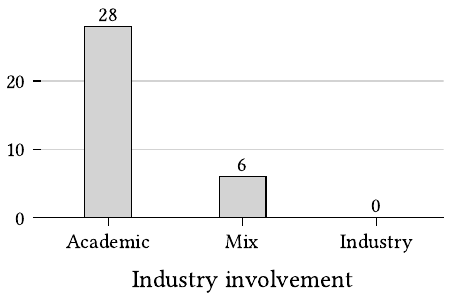}
    \caption{Distribution of papers according to the type of industry involvement.\label{fig:plot_industry}}
\end{figure}

\highlight{\textbf{Industry Involvement:}
The presence of industrial researchers among the authors is still scarce.}

\subsection{Tool Provision}\label{sec:tool}
From the set of primary studies, only very few papers (\papercount{2}) provide available tools to address the mobile energy consumption involving artificial intelligence.
\textit{GreenAdvisor}, a Java-based tool, is made available by Aggarwal~\etal~\citeP{aggarwal2015greenadvisor}. It advises Android application developers on change in energy consumption of their app with change in app code, based on system call tracing\footnote{https://github.com/kaggarwal/GreenAdvisor}.
Nawrocki~\etal~\citeP{nawrocki2019adaptable} provide a TypeScript library for Mobile Cloud Computing code offloading with Machine Learning\footnote{https://github.com/Hoobie/ml-offloading}.

\highlight{\textbf{Tool Provision:}
Although many studies provide solutions to address mobile energy consumption involving AI, only a small portion of them make the solution-based tools readily available online.}

\section{Discussion} \label{sec:discussion}
A clear picture emerges from the analysis of publishing trends on mobile energy consumption involving AI.
As shown in Section \ref{sec:year}, a recent increase in the number of papers published on the topic can be observed. We can speculate on a few reasons for this. AI in general has seen a great increase in interest over the past few years, driven by major technical developments in deep learning, such as shown by Mnih~\etal in reinforcement learning \cite{mnih2015human}, Goodfellow~\etal in the development of the Generative Adversarial Network (GAN)~\cite{goodfellow2020generative}, and the recent development of models like Minerva, which can solve undergraduate-level mathematical and engineering problems~\cite{lewkowycz2022solving}. 
Secondly, AI on mobile devices has become more feasible both due to generic improvements to hardware and software, but especially due to the inclusion of so-called accelerators, such as Apple's Neural Engine \cite{banerjee2018microarchitectural} and Google's TPU~\cite{gupta_2021}.
Finally, energy consumption and efficiency are becoming increasingly important across all sectors due to ever-escalating global warming. Therefore, as this issue becomes more pressing, one can expect an increase in the interest in energy efficiency improvements in technology. 


Considering the explored topics, in general, a fair body of research on artificial intelligence and mobile energy use exists. However, the works in this body are often highly specific to a specific technology, such as \textit{Federated Learning}. However, even Federated Learning itself is partly an accounting trick: although the model is only trained partially on each local device, thereby reducing energy consumption per training device, this energy use is simply spread over multiple devices. Accounting for the communication energy required by each device to get the model from the server and send it back, a net training energy consumption calculation might turn out poorly for federated learning.

We also noticed that having access to up-to-date benchmarks is a major challenge in this field. For example, Deng~\etal~\citeP{deng2022making} was published in 2022 but uses baselines from 2016 and 2018. This can be an indicator that the field is still evolving rapidly, or that it is more difficult to publish such studies when involving AI in Green Mobile Software.

Regarding the two faces of AI and its involvement in mobile energy consumption, it can be noted that the number of papers in this review on the topic of using AI to reduce energy consumption was twice as high as the number of papers on reducing the energy consumption of AI itself. All this work cited or was cited by Anh~\etal \citeP{anh2019efficient} or McIntosh ~\etal~\citeP{mcintosh2019can}, both of which were published years ago already. In general, a lack of broad research on this topic can be observed, except for papers addressing Federated Learning energy use. However, as discussed earlier in the discussion, some of the benefits of Federated Learning can be seen as an accounting trick. 
The body of work focused on using AI to reduce energy consumption is extensive, but it remains confined to a relatively limited range of research topics. Networking optimization and offloading techniques appear to be prominent areas with intense competition driving innovative approaches. However, other areas, such as quantifying the impact of code changes in mobile apps or optimizing video playback, have received less exploration. Future research in these areas could contribute significantly to further reducing mobile energy consumption, which is crucial given the projected growth in the number of mobile devices in the coming years.


Thanks to the findings of this review, it is evident that the systematic provision of tools addressing AI and Green Mobile Software is far from optimal. While many studies propose solutions to address mobile energy consumption using AI, only a small portion of them make their solution-based tools readily available online. This observation suggests two potential explanations: (i) the rapid evolution of research in this field, which leads to quickly outdated results and renders tools less meaningful, or (ii) an immaturity in the research field that requires stronger empirical support as a foundation for the development of tools.

Finally, from our results, the literature seems to be mainly involving the academic community, as the majority of the primary studies presents an academic authorship only. Again, although there is a predominance of solutions papers, only a very few industrial researchers are authors of papers dealing with AI in Green Mobile Software. This can be related to the fact that few tools are made available online.
We argue that, especially as mobile devices are getting more and more prevalent, to have a real impact on the energy efficiency of mobile software, and to update the related practises, we need to promote the involvement of the mobile software industry in this field.

\section{Threats to Validity} \label{sec:threats}
In this section, we discuss the threats to validity of our research. To ensure the quality of the results, we established a well-defined research protocol before performing the data collection. In addition, we adhered to a set of well-accepted snowballing guidelines for literature reviews \cite{wohlin2014guidelines, Kitchenham2004, petersen2015guidelines, mayring2004qualitative}. To lower potential sources of bias, crucial considerations that emerged during the research were discussed among the authors. Despite adhering to a systematic literature review approach, potential threats to validity remain. The remainder of this section addresses four types of validity: internal, external, construct, and conclusion validity.

\subsubsection{Internal validity}
To address internal validity threats, we applied a systematic snowballing and open coding process. We used existing guidelines, such as \cite{wohlin2014guidelines, mayring2004qualitative, maurer2008card}, to avoid inventing new methods and prevent us from making non-obvious but consequential methodological mistakes. However, even though the choices were discussed among the authors, some bias in paper selection and coding might remain.

\subsubsection{External validity}
The main threat to external validity of this study is the lack of representativeness of the papers considered. We think this threat can manifest in three main ways: 1) The primary studies, found using an academic search engine, are not representative of the topic we attempt to study. We mitigate this threat by using Google Scholar, which searches very broadly among publishers of literature. 2) The set of studies is incomplete. We mitigate this threat by performing snowballing, both forward and backward. However, no snowballing process can guarantee a complete set of literature, and only one iteration was performed.
3) The potential non-relevance of the papers. We have mitigated this threat by using in- and exclusion criteria that address quality, such as the requirement of being peer-reviewed. 

\subsubsection{Construct validity}

To ensure the relevance of the primary papers in addressing our research questions, we applied meticulously crafted inclusion and exclusion criteria. Subsequently, bidirectional snowballing was employed to expand the set of relevant primary papers. However, we conducted only one iteration of snowballing, and further iterations could have potentially resulted in the inclusion of additional papers.

\subsubsection{Conclusion validity}
Sources of bias from our analyses are addressed by following a strict and clearly defined process based on public guidelines~\cite{wohlin2014guidelines, maurer2008card, mayring2004qualitative}.
Lastly, we documented all the data throughout the whole review process and made them available through a replication package for reproducibility and replicability purposes (cf. Section 1).

\section{Related work} \label{sec:related}
Several secondary studies have examined related topics. We briefly discuss them here.

In a recent systematic review~\cite{SLRGreenAI}, the focus is on Green AI, specifically the energy consumption of AI itself. Interestingly, the authors found that studies on mobile computing comprise only a small fraction (around 4\%) of the Green AI literature. Our review goes beyond mobile computing to explore how AI can reduce mobile energy consumption.

Measurement methods for mobile energy are discussed by Khan \etal~\cite{khan2021measuring}, where 21 techniques (hardware and software-based) are described and compared. While previous work~\cite{ahmad2015review} covers similar ground, it was published in 2015, analysing a different period. Some techniques mentioned in these papers have been used to measure energy consumption in certain AI applications, but they may not be applicable to other AI systems like federated learning, which we focus on here.

General energy management for mobile devices is addressed in~\cite{pasricha2020survey}. Other reviews cover specific areas like video streaming~\cite{deshpande4exploring} and processing unit management~\cite{kim2018survey}. In contrast, our work specifically examines AI's role in energy management within mobile software.

Studies on energy consumption in mobile cloud computing (MCC) \cite{somula2019research} and \cite{parajuli2020recent} compare MCC frameworks and review multimedia application papers, respectively. However, our focus is exclusively on works that discuss AI applications.

Shi \etal~\cite{shi2021mobile} explore mobile edge AI techniques, including model compression, federated learning, and offloading. While our work intersects with this topic, we emphasize using AI to save energy rather than reducing AI's energy use.

Federated learning is extensively surveyed in ~\cite{shi2022towards}, with a focus on communication costs in 5G networks. Our work intersects partially, considering the use of these AI techniques to save energy.

\section{Conclusion} \label{sec:conclusion}
In this paper, we provide an overview of the literature on mobile energy consumption and artificial intelligence. Our research questions reflect on the publication trends in this area and the characteristics of 34 papers. We describe two main branches in the literature: 1) papers looking at the energy consumption of mobile AI applications and 2) papers focusing on applying AI to reduce mobile energy consumption.

We identify groups of papers that consider similar topics or use similar techniques. We pinpoint main research directions, such as offloading and networking optimization to save energy on mobile devices and the analysis of the energy consumption of federated learning. However, other areas, such as approximation computing, have been less investigated. 

For researchers, this paper provides an overview of this research area and it points to promising directions for future research. It is also relevant for stakeholders in the mobile computing industry, as we identify potential solutions that arise from the deployment of artificial intelligence models in mobile apps. It also helps in identifying areas where further research and investment are needed.

\balance
\bibliography{mybib}

\nociteP{*}
\bibliographystyleP{ieeetr}
\bibliographyP{primaries}

\end{document}